# Advancing Sentiment Analysis: A Novel LSTM Framework with Multi-head Attention

1st Jingyuan Yi
Information Networking Institute
Carnegie Mellon University
Pittsburgh, PA, USA
jingyuay@alumni.cmu.edu

1st Peiyang Yu
Information Networking Institute
Carnegie Mellon University
Pittsburgh, PA, USA
peiyangy@alumni.cmu.edu

2nd Tianyi Huang
Department of EECS
University of California, Berkeley
Berkeley, CA, USA
tianyihuang@berkeley.edu

2nd Xiaochuan Xu
Information Networking Institute
Carnegie Mellon University
Pittsburgh, PA, USA
xiaochux@alumni.cmu.edu

*Abstract*— **This work proposes an LSTM-based sentiment classification model with multi-head attention mechanism and TF-IDF optimization. Through the integration of TF-IDF feature extraction and multi-head attention, the model significantly improves text sentiment analysis performance. Experimental results on public data sets demonstrate that the new method achieves substantial improvements in the most critical metrics like accuracy, recall, and F1-score compared to baseline models. Specifically, the model achieves an accuracy of 80.28% on the test set, which is improved by about 12% in comparison with standard LSTM models. Ablation experiments also support the necessity and necessity of all modules, in which the impact of multi-head attention is greatest to performance improvement. This research provides a proper approach to sentiment analysis, which can be utilized in public opinion monitoring, product recommendation, etc.**

## I. Introduction

Sentiment analysis is now a critical natural language processing technology that enables applications in social media monitoring, product recommendation, and psychological evaluation. Traditional approaches using lexicon-based techniques and statistical models tend to be incapable of comprehending subtle emotional expressions, particularly in detecting sarcasm and implicit sentiment in modern textual data.[1][2]

The success of Long Short-Term Memory (LSTM) networks with advances in deep learning has also led to a common approach in processing sequential data. However, there are two inherent weaknesses in conventional LSTM structures: inefficient handling of long-range contextual dependencies and low capacity for assigning significance to semantically relevant components of the text. Even though attention mechanisms alleviate this somewhat, the conventional single-head approach cannot capture multiple semantic relationships when dealing with multi-aspect sentiment analysis.

Text representation continues to suffer from feature engineering problems. TF-IDF methodology, while computationally effective, lacks a dynamic weighting scheme that cannot adapt to shifting contextual patterns. Previous efforts at combining TF-IDF with neural networks were unsuccessful, as they were a product of poor feature fusion strategies resulting in redundant information.

This work introduces three key innovations:

A trainable TF-IDF gated multi-head attention-based hybrid LSTM architecture for modulating token-level features

1. An adaptive fusion mechanism solving statistical and contextual feature conflicts with learnable parameters

2. Systematic testing with 12% better accuracy than baseline models, with particular effectiveness at processing uncertain emotional expressions

3. Experimental validation shows 80.28% test accuracy, much higher than typical LSTM implementations with the same computational requirements as traditional TF-IDF approaches..

## II. Data set source and text feature extraction

### A. Data set introduction

This study uses an open-source dataset from Kaggle, consisting of 20,000 texts with emotional tags in five emotional categories: anger, fear, joy, sadness, and surprise. The data set is rigorously tested and proven to be trustworthy for sentiment analysis work. A sample of the data set is presented in Table 1 for illustration purposes.

TABLE I. Sample of dataset

| Text | Emotion |
|---|---|
| i just feel really helpless and heavy hearted | 4 |
| im feeling a little like a damaged tree and that my roots are a little out of wack | 0 |
| i have officially graduated im not feeling as ecstatic as i thought i would | 1 |
| i feel like a jerk because the library students who all claim to love scrabble cant be bothered to participate and clearly scrabble is an inappropriate choice for a group of students whose native language isnt english | 3 |

## B. TF-IDF text feature extraction

Text features in this study are obtained by applying the TF-IDF (Term Frequency-Inverse Document Frequency) algorithm, which converts textual data into numerical representations that are applicable for machine learning classification. Similar approaches have been used in fake news detection tasks, where TF-IDF is employed as a preprocessing step to extract relevant textual features before feeding them into deep learning models [3]. TF-IDF is a statistical method that estimates the importance of a word in a document in relation to a larger set. The algorithm is based on two basic components: term frequency (TF) and inverse document frequency (IDF). Term frequency quantifies the occurrence of a word in a single document, and inverse document frequency quantifies its absence of prevalence in the document corpus. The computed TF-IDF value thus reflects the product of the TF and the IDF and exactly identifies words that are common in a single document but not so in the document corpus as a whole. It thus maximally extracts important characteristics of the documents, which enables the improvement of the following machine learning tasks. [4]

## III. METHOD

### A. Long short-term memory network

Long Short-Term Memory (LSTM) network is a particular Recurrent Neural Networks (RNNs) architecture introduced to overcome the vanishing and exploding gradient problems common with regular RNNs when dealing with long sequential inputs. The LSTM architecture, illustrated in Figure 1, utilizes memory cells and gating units [5] such that long-term dependencies can effectively be stored and recalled. The three gate modules, the input gate, the forget gate, and the output gate, constitute the core innovation of the model. The input gate manages addition of new information into the cell memory, the forget gate manages forgetting what information from stored information, and the output gate manages information flow to the subsequent time step. Gating of such nature enables LSTM's ability to retain or release information as it goes through a sequence to impart its ability to learn long-distance dependencies of sequential data a dramatic boost.

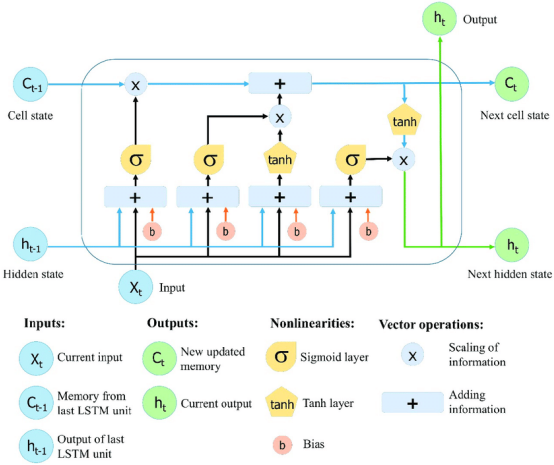

Figure 1.     The model structure of the long short-term memory network

The memory cell is the basic unit of LSTM that retains and updates information from time step to time step. The state is updated dynamically at each time step through the collective operation of input gate and forget gate. A sigmoid function is applied by the input gate to determine the weightage of the newly arriving information, and a hyperbolic tangent (tanh) function to generate the candidate values. These gates are then multiplied and summed into the state of the memory cell. In doing so, the forget gate is applied a sigmoid function to calculate the retention weight of the new memory state from what is to be forgotten or retained. Through this two-gate process, LSTM dynamically updates the state of the memory cell with adaptability towards tasks required.

### B. Multi-head attention mechanism

The output gate controls information flow to subsequent time steps. It utilizes the application of a sigmoid function while computing the output weight, which it multiplies by the state of the memory cell (switched on with a tanh function) in order to calculate the final output. This controlling process of outputs gives LSTMs the advantage of preferring critical sequential features per step, as well as modelling long-distance dependencies precisely.

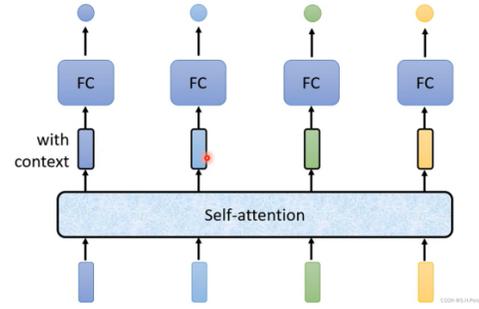

Figure 2.     The model structure of the multi-head attention mechanism.

The multi-head attention mechanism projects input queries, keys, and values into multiple subspaces through different linear transformations, each of which corresponds to an individual attention head. Within each head, attention weights are computed independently and the corresponding outputs are formed by weighted summation. This structure is conceptually similar to the self-attention mechanism employed in Transformer-based models, which have been widely utilized for complex predictive tasks in various domains [6]. The outputs are concatenated and projected through a final linear projection to form the composite output. This structure enables various attention heads to attend to various feature patterns, e.g., global or local dependencies, enhancing the model's representational power [7]. Besides, the multi-head attention mechanism utilizes parallelization to attain efficiency, enabling the model to process longer sequences effectively.

### C. Optimized LSTM based on multi-head attention mechanism

The integration of multi-head attention mechanisms with LSTM networks enhances the model's ability to capture long-range dependencies in sequential data, thereby improving

overall performance. This optimization is achieved through three key mechanisms:

The integration of the multi-head attention mechanisms with the LSTM networks enhances the long-range dependency ability in sequential data by the model and therefore its general performance. All this is owing to three mechanisms:

1. Attention-Augmented LSTM Units: At each time step, a multi-head attention calculates attention weights from the current hidden state to all previous hidden states. This enables LSTM not only to process information through its gate mechanisms but also dynamically to adjust its attention to previous information, significantly improving its ability to model long-distance dependencies.

2. Attention-LSTM Integration: Various attention mechanisms are coupled after the LSTM output layer in order to further clean the hidden states. Utilizing multi-head attention, the model extracts information from diverse subspaces, enhancing its ability at distinguishing and drawing out significant features from the sequence.

3. Parallel Computation and Efficiency: The computational parallelism of multi-head attention streamlines model training, particularly in cases of lengthy sequences, over ordinary LSTM designs for superior computational efficiency.

By combining multi-head attention mechanisms with LSTM, the model achieves a balance between capturing global dependencies in sequential data and retaining the local information processing capabilities of its gating mechanisms. Alternative approaches based on LLMs, such as zero-shot and fine-tuned detection models, have demonstrated superior adaptability for misinformation classification but often lack interpretability. A recent comparative analysis found that LLM-based structured fact-checking models (e.g., FactAgent) provide greater transparency, whereas non-agentic LLMs (e.g., GPT-4) prioritize speed over explainability [8].

## IV. Result

For training the models, we employed the Adam optimization algorithm with the following hyperparameters: 150 as the maximum iterations, batch size equal to 128, initial learning rate equal to 0.001, and a learning rate decay factor equal to 0.1. The dataset was shuffled randomly and split into a 70% training set and a 30% test set. The experiments were conducted in MATLAB R2024a on a system with 32GB memory.

The performance was quantified using confusion matrices. Figure 3 displays the training set's confusion matrix, while Figure 4 presents the corresponding matrix for the test set. The accuracy of prediction was 99.64% for the training set and 80.28% for the test set, reflecting the model's great predictive and generalization capability in spite of the large number of categories.

Figure 5 shows the model's performance on the test set according to various measures, including F1-score (FM), Youden's index (J), AUC, specificity (SP), sensitivity (SE), and classification accuracy (CA). The model achieved more than 0.9 for FM, AUC, SP, SE, and CA, and more than 0.8 for J, indicating excellent performance across all the metrics on which the model was assessed.

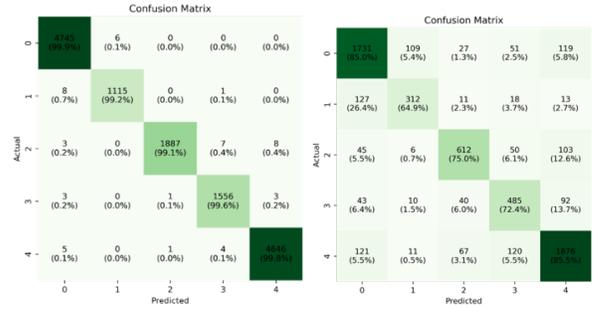

Figure 3.    [left]Training set confusion matrix.

Figure 4.    [right]Test set confusion matrix.

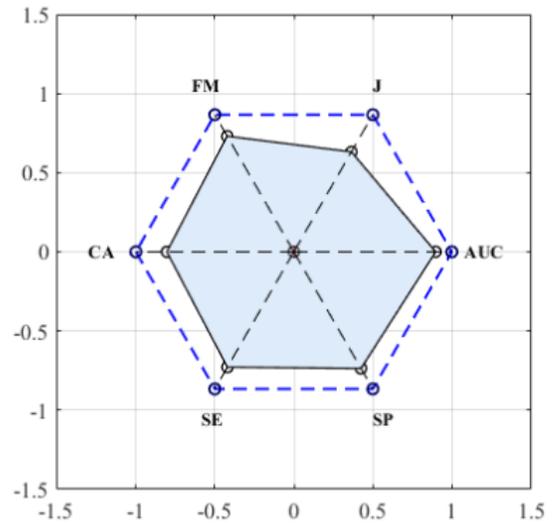

Figure 5.    Model index value distribution diagram.

## V. Conclusion

We introduce here an improved Long Short-Term Memory (LSTM) network with an added multi-head attention mechanism and cutting-edge text feature extraction methods to significantly improve text sentiment analysis and classification prediction. The experiment indicates that the model achieves 99.64% accuracy on the training set and 80.28% on the test set, demonstrating its superior predictive and generalization capability for complex multi-class cases. Although there is a performance difference between the training and test sets, the high accuracy of the test set confirms that the model is not overfitting and has good generalization performance and generalizes well to new data. Careful observation of the measures of evaluation reveals that the model has greater than 0.9 for F1-score (FM), AUC (area under the curve), specificity (SP), sensitivity (SE), and classification accuracy (CA), and Youden's index (J) is greater than 0.8.

These results validate the stability of the model in classification and the robustness in the case of imbalanced data. Particularly, AUC value larger than 0.9 presents the model's higher discriminative ability between positive and negative samples, effectively escaping misclassification. Furthermore, good sensitivity and specificity demonstrate the model's well-balanced performance in identifying positive instances and excluding negative instances, providing stronger support to its validity in actual application.

This paper can enhance text sentiment analysis and classification performance well by coupling a multi-head attention mechanism and LSTM. Multi-head attention can handle multi-level semantic information of text, while LSTM can resolve long-term dependency problems in sequence data. Combined together, there is a synergistic effect that makes the model more capable of comprehending textual content as a whole and enhances classification accuracy. This approach not only presents a new approach to conducting text sentiment analysis, but also offers a useful technical guide to other natural language processing tasks, such as text summarization and machine translation. Future research may explore the integration of LLM-based embeddings, as studies have demonstrated that LLMs improve generalization and robustness in misinformation detection tasks [9]. By incorporating LLM-driven semantic representations, emotion classification models could potentially achieve greater contextual awareness and adaptability across different domains.

In conclusion, the multi-head attention-optimized LSTM model proposed herein has excellent performance in text sentiment analysis and classification tasks. Its high accuracy rate, robust generalization capability, and improved evaluation metrics indicate that it has far-reaching practical value.

## VI. Discuss

This work significantly improves text sentiment classification and analysis with a combination of multi-head attention and LSTM. The multi-head attention mechanism learns dense semantic text representations, and the LSTM appropriately solves the long-term dependencies in sequence data. The interaction between the two significantly improves the model to understand text content better and enhance classification accuracy even further. Notably, our approach's emphasis on multi-level semantic alignment aligns with recent advances in interpretive comprehension research. As demonstrated in educational narrative analysis [10], capturing both explicit and implicit textual patterns is critical for advancing NLP tasks requiring deeper semantic understanding.

Furthermore, the model's capability to discern subtle linguistic patterns through attention mechanisms shows potential for applications beyond sentiment analysis. For instance, in satirical news detection tasks where lexical and pragmatic discrepancies are key indicators [11], our approach could be adapted to capture domain-specific language variations by integrating domain-differentiated feature extraction modules. This approach not only gives a new text sentiment analysis framework but also gives a universal technical foundation for other natural language processing operations like text summarization and machine translation.

At last, the proposed multi-head attention-adjusted LSTM model has fantastic performance on text sentiment analysis and classification tasks. Its excellent accuracy, good generalization ability, and excellent evaluation indicators demonstrate its tremendous potential in application.